\documentclass{article}

     \usepackage[preprint]{neurips_2019}

\usepackage[utf8]{inputenc} 
\usepackage[T1]{fontenc}    
\usepackage{hyperref}       
\usepackage{url}            
\usepackage{booktabs}       
\usepackage{amsfonts}       
\usepackage{nicefrac}       
\usepackage{microtype}      

\usepackage{graphicx}
\usepackage{amsbsy}

\usepackage{calligra}
\usepackage{eucal}
\usepackage{adjustbox}

\usepackage[toc,page]{appendix}
\usepackage{wrapfig}

\usepackage{hyperref}

\title{Cosine similarity-based adversarial process}

\author{%
  Hee-Soo Heo \\
  School of Computer Science\\
  University of Seoul, South Korea\\
  \And
  Jee-weon Jung \\
  School of Computer Science\\
  University of Seoul, South Korea\\
  \And
  Hye-jin Shim \\
  School of Computer Science\\
  University of Seoul, South Korea\\
  \And
  IL-Ho Yang \\
  School of Computer Science\\
  University of Seoul, South Korea\\
  \And
  Ha-Jin Yu \thanks{Corresponding author} \\
  School of Computer Science\\
  University of Seoul, South Korea\\
}

\begin{document}

\maketitle

\begin{abstract}
An adversarial process between two deep neural networks is a promising approach to train a robust model.
In this paper, we propose an adversarial process using cosine similarity, whereas conventional adversarial processes are based on inverted categorical cross entropy (CCE).
When used for training an identification model, the adversarial process induces the competition of two discriminative models; one for a primary task such as speaker identification or image recognition, the other one for a subsidiary task such as channel identification or domain identification. 
In particular, the adversarial process degrades the performance of the subsidiary model by eliminating the subsidiary information in the input which, in assumption, may degrade the performance of the primary model. 
The conventional adversarial processes maximize the CCE of the subsidiary model to degrade the performance. 
We have studied a framework for training robust discriminative models by eliminating channel or domain information (subsidiary information) by applying such an adversarial process. 
However, we found through experiments that using the process of maximizing the CCE does not guarantee the performance degradation of the subsidiary model.
In the proposed adversarial process using cosine similarity, on the contrary, the performance of the subsidiary model can be degraded more efficiently by searching feature space orthogonal to the subsidiary model. 
The experiments on speaker identification and image recognition show that we found features that make the outputs of the subsidiary models independent of the input, and the performances of the primary models are improved.

\end{abstract}

\section{Introduction}

A speech waveform can have many kinds of information such as the phonetic information, speaker identity, age, dialect, health condition, channel, environment noises, etc. 
If our task is to identify one of those information, then other information can confuse the result. 
For example, if our task is to identify speakers, channel information can degrade the performance greatly. 
It is hard for the machine to discriminate between the channel differences from the speaker differences. 
We call the annoying information "subsidiary information." 
The same problem applies to almost pattern recognition tasks. 
There have been numerous researches for removing the disturbing information for ages. 
Recently a method is proposed that using discriminative adversarial network (DAN).

\begin{figure}[ht]
\begin{center}
\centerline{\includegraphics[width=0.6\columnwidth]{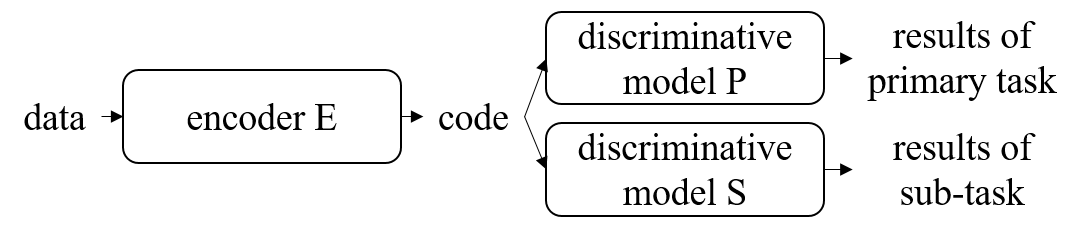}}
\caption{Overall architecture of the DAN framework which includes primary task and subsidiary task (sub-task).}
\label{fig1}
\end{center}
\vskip -0.2in
\end{figure}

DAN is a framework designed to train discriminative models by using an adversarial process, unlike the generative adversarial network (GAN, \cite{goodfellow2014generative}) used for generative models (\cite{ADDA}, \cite{GRL2015}, \cite{GRL2016}, \cite{yu2017adversarial}). 
This framework is composed of three components: an encoding network E, a discriminative model P for the primary task, and a discriminative model S for the subsidiary task. 
The primary task is the main goal of our research. 
In contrast, the subsidiary task classifies the subsidiary information that can hinder and degrade the primary task. 
For example, if digit recognition is the primary task, domain recognition could be a subsidiary task; if speaker identification is the primary task, channel identification could be a subsidiary task. 
The channel information and domain information included in the input data are well known factors that degrade the performance of the corresponding primary task.
E encodes the input data into a fixed-dimensional code; E can be composed of convolutional layers as well as fully connected layers. 
P and S takes the code from E as input and perform the classification based on the primary task and subsidiary task, respectively. 
Fig. 1 shows the overall architecture of the DAN framework. 
The objective of the DAN framework is to degrade the performance of S by training E so that subsidiary information (domain or channel information) contained in the code is eliminated and the performance of P increases consequently because the subsidiary information have confused P.  

The training procedure to achieve this objective is as follows. 
The first stage is to train E and P to perform the primary task.
In particular, we can use loss $\mathcal{L}_{P}$ defined in Eq. (1) based on categorical cross entropy (CCE) to train E and P. 
\begin{equation}
\mathcal{L}_{P} = - \sum\limits_{ (\boldsymbol{x},y^P) \in X}log\frac{exp(W_{y^P}^{P}E(\boldsymbol{x})+b_{y^P}^{P})}{\Sigma_{j=1}^{N^{P}}exp(W_{j}^{P}E(\boldsymbol{x})+b_{j}^{P})},
\end{equation}
\noindent where the superscript $(\cdot)^P$ means that each variable is related to the primary task, $X$ is the training set, $\boldsymbol{x}$ and $y^P$ are the input data and the corresponding label index, respectively, for the primary task, $W^{P}$ and $\boldsymbol{b}^{P}$ are the parameters of the output layer of the discriminative model P, $W_{i}^{P}$ is the $i'th$ row of matrix $W^P$, $b_{i}^{P}$ is the $i'th$ element of vector $\boldsymbol{b}^{P}$, $N^P$ is the number of classes in the primary task, and $E(\cdot)$ is the output of the encoder E. 
Eq. (1) is defined based on the assumption that P consists of one output layer. 
Thus, $W^P$ is a matrix of size $N^P \times D^E$ and $b$ is a vector of length $N^P$, where $D^E$ is the dimension of the output of E. 

The second stage is to train S to perform the subsidiary task. 
At this stage, S takes the codes from fixed E as input, and is trained by using $\mathcal{L}_{S}$ defined in Eq. (2).
\begin{equation}
\mathcal{L}_{S} = - \sum\limits_{ (\boldsymbol{x},y^S) \in X}log\frac{exp(W_{y^S}^{S}E(\boldsymbol{x})+b_{y^S}^{S})}{\Sigma_{j=1}^{N^{S}}exp(W_{j}^{S}E(\boldsymbol{x})+b_{j}^{S})},
\end{equation}
\noindent where the superscript $(\cdot)^S$ means that each variable is related to the subsidiary task, $y^{S}$ is the label for the subsidiary task, $N^S$ is the number of classes in the subsidiary task, and $W^{S}$ and $\boldsymbol{b}^{S}$ are the parameters of the output layer of the discriminative model S.
The above Eq. (2) is defined on the assumption that S consists of one output layer and also based on CCE. 
The subsidiary information included in the codes can be grasped in this stage. 

The third stage is to apply the adversarial process based on the trained S to eliminate the subsidiary information included in the codes. 
At this stage, various types of losses are defined and applied to the adversarial process; most losses are defined by reversing the loss defined by Eq. (2), as shown in Eq. (3).
\begin{equation}
\mathcal{L}_{adv} = -\mathcal{L}_{S}.
\end{equation}
It is important to note that only E is trained by $\mathcal{L}_{adv}$ in the third stage. 
Therefore, the encoder E is trained to maximize $\mathcal{L}_{S}$ in the third stage while S is trained to minimize $\mathcal{L}_{S}$ in the second stage. 
Generally, this adversarial process is implemented by reversing the CCE or using the gradient reversal layer (GRL) (\cite{GRL2016}, \cite{GRL2015}). 

\begin{figure}[ht]
\begin{center}
\centerline{\includegraphics[width=0.9\columnwidth]{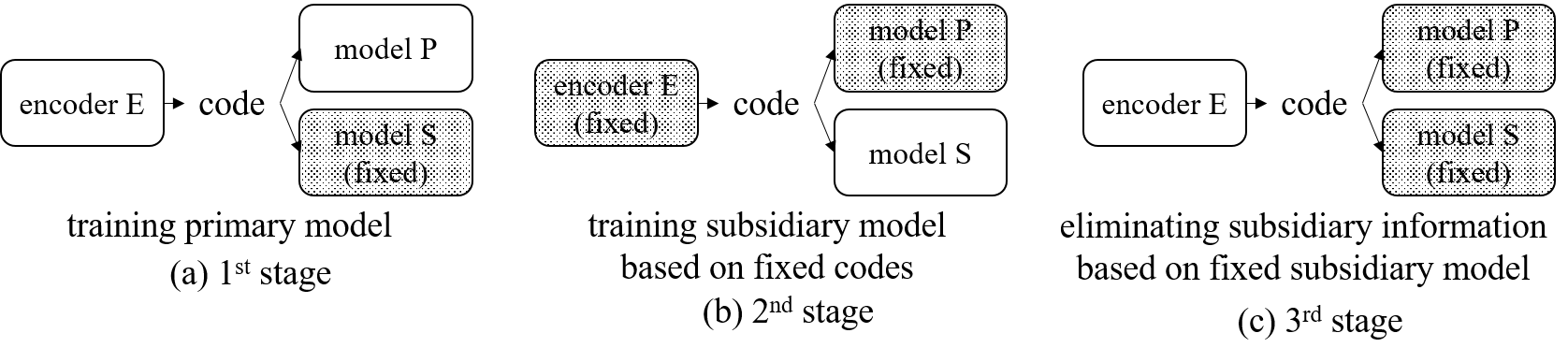}}
\caption{Illustration of the training procedure of the DAN framework. (a) First stage: training the encoder E and the model P for primary task; (b) second stage: training the model S (subsidiary parameters) based on fixed codes; (c) third stage: training the encoder E to eliminate subsidiary information based on the fixed model S.}
\label{fig1}
\end{center}
\vskip -0.2in
\end{figure}

The three stages explained above can be summarized via Fig. 2-(a), (b), and (c), respectively.
It was shown that the subsidiary information included in the codes could be eliminated and the primary model could be made more robust to subsidiary information by means of the DAN framework that repeats from the first to the third stage.

\section{Limitation of the conventional adversarial process}
In the studies using the DAN framework introduced, the adversarial processes are implemented as the process of maximizing the CCE (as shown in Eq. (3)). 
A limitation of the conventional adversarial processes (\cite{ADDA}, \cite{GRL2015}, \cite{GRL2016}) is that the process of maximizing the CCE does not guarantee the removal of the subsidiary information which leads to the performance degradation of S.  
Assuming S that identifies four classes and was trained by minimizing the CCE, we can expect the following outputs from S; $S(\boldsymbol{x}_{1})=[1,0,0,0]^T$ when the $\boldsymbol{x}_1$ is from the first class, or $S(\boldsymbol{x}_{2})=[0,0,1,0]^T$ when the $\boldsymbol{x}_2$ is from the third class. 
Note that the output of S is calculated using softmax function and is a vector of length four where each element is related to the corresponding class. 
Therefore, we can classify each data to the corresponding class by applying argmax function to the outputs of S (\cite{deepbook}). 
On the other hand, if S was trained to maximize the CCE by the conventional adversarial process, S may output the followings; $S(\boldsymbol{x}_{1})=[0,0.2,0.3,0.5]^T$ or $S(\boldsymbol{x}_{1})=[0,0.7,0.2,0.1]^T$ when the $\boldsymbol{x}_1$ is from the first class. 
Note that the output that maximizes the CCE for the fixed input $\boldsymbol{x}_1$ is not unique. 
Analyzing the outputs of S for $\boldsymbol{x}_1$, it can be interpreted that S can not perform identification since the element corresponding to the first class has a value of 0. 
However, $\boldsymbol{x}_1$ still can be identified as the first class by applying argmin function to the outputs of S. 
As mentioned earlier, the purpose of applying the adversarial process is to eliminate the subsidiary information by degrading the performance of S. 
However, we found that the conventional adversarial process of maximizing the CCE does not degrade the performance of S. 
This is because the process of maximizing the CCE minimizes the term $W_{y^S}^{S}E(\boldsymbol{x})$ in the Eq. (2).
This process can be interpreted as the training process of the classification model that maximizes the negative correlation between the data $\boldsymbol{x}$ and the corresponding class label $y$. 
The empirical evidence about this interpretation will be presented in Section 4.1.
Therefore, we propose a novel adversarial process using cosine similarity to efficiently degrade the performance of S. 
Specifically, we propose a discriminative model training framework using a novel adversarial process based on cosine similarity (cosine adversarial network, CAN) to overcome the limitations of the conventional DAN framework.

\section{Proposed cosine similarity-based adversarial network}

In this section, we describe the proposed CAN framework. 
As explained in the previous section, the conventional adversarial processes may not be successful in degrading the performance of S and eliminating subsidiary information efficiently. 
Therefore, we designed a novel adversarial process that makes the output of the subsidiary network independent to the input using the concept of rotation to degrade the performance of S more efficiently and to guarantee the operation of P.  
Assume that E is trained to output zero vector ($E(\boldsymbol{x})=\boldsymbol{0}$ for any $\boldsymbol{x}$).
This will degrade the performance of S significantly; however, the operation of P, the main purpose of the system, will not also be guaranteed.
This is why we adopted the concept of rotation that can only change the direction of the code while keeping the scale of the code. 
Fig. 3 shows an example of the operation of S and the proposed rotation in two-dimensional code space. 

\begin{figure*}[ht]
\begin{center}
\centerline{\includegraphics[width=0.9\columnwidth,trim=4 4 4 4,clip]{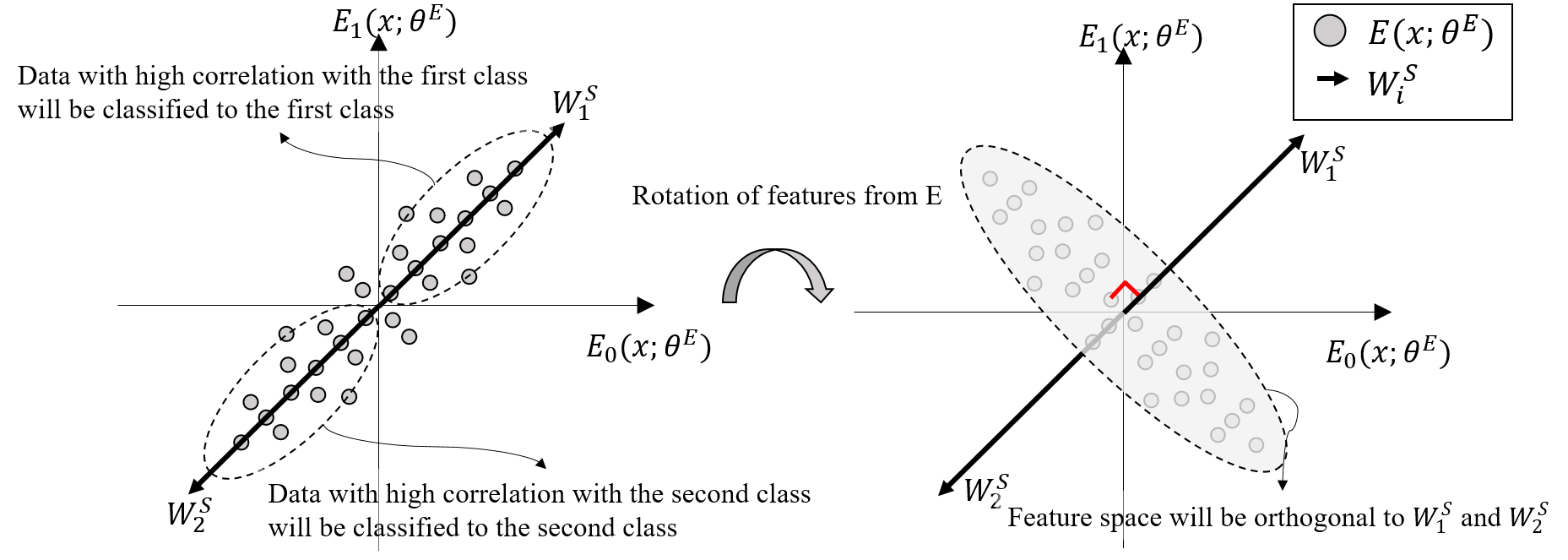}}
\caption{The rotation of codes from E by applying the proposed cosine similarity-based adversarial process.}
\label{fig1}
\end{center}
\vskip -0.2in
\end{figure*}

In this figure, the code space is defined by E, and the subsidiary task is performed based on weight parameters $W_{i}^{S}$ of S. 
Therefore, we can expect that the performance of S decrease, if the code space is rotated to be orthogonal to $W_{i}^{S}$ of S. 
With this expectation, we define the adversarial loss to train E based on cosine similarity as follow:
\begin{equation}
\mathcal{L}_{cos} = \sum\limits_{ \boldsymbol{x} \in X}\sum\limits_{j=1}^{N^{S}} CS( W_{j}^{S},E(\boldsymbol{x}))^{2},
\end{equation}
\noindent where $CS(\cdot, \cdot)$ is the cosine similarity between two vectors, $W_{j}^{S}$ is $j'th$ row of weight matrix from the output layer of the discriminative model S, and $E(\cdot)$ is the output of the encoder E.
Eq. (4) is defined based on the assumption that S consists of one output layer. 
This loss function sets the absolute values of the cosine similarity between the codes from E and all rows of the weight matrix $W^{S}$ to zero.
Thus, rotating the codes almost orthogonal to the weight parameters of S. 
To avoid the problem of non-division and improve the efficiency of outlier training, the square function is used instead of the absolute values. 
The objective of $\mathcal{L}_{cos}$ is to eliminate the subsidiary information by making the codes from E independent of the subsidiary class label $\ y^{S} $. 
The dependency between the codes and the label $\ y^{S} $ can be modeled by the weight parameters of S. 
Underlying hypothesis in this context is that making the codes orthogonal to the weight parameters of S will make the codes independent of the class label $\ y^{S} $, thereby eliminating the subsidiary information contained in the codes. 
Note that we did not use the label $y^S$ to define $\mathcal{L}_{cos}$.

The effect of our proposed framework CAN on the S is different from the conventional adversarial process. 
Assuming that the proposed framework minimizes the loss defined in Eq. (4) to zero, the classification of the subsidiary task will be impossible. 
The output of the S, defined by the weights matrix $\ W^{S}$ and bias term $\ b^{S}$, is calculated as $\ \boldsymbol{o}^S = W^{S}E(\boldsymbol{x}) + b^{S}  $ as in Eq. (2). 
In addition, the term $\ W^{S} E(\boldsymbol{x})  $ will be zero when the loss function defined by Eq. (4) is zero. 
Thus, the output is only dependent on the bias term, which is a constant term; consequently, the S has no discriminative power for any input $ \boldsymbol{x} $.

\section{Experiments}
\label{app}

In this section, we show experiments using the proposed adversarial processes. 
In order to apply the CAN framework, speaker identification and digit recognition are set as the primary tasks; in addition, channel information and domain information are set as the subsidiary information for each task, respectively. 
All deep neural networks used for these experiments were implemented using Keras (\cite{keras}) with a TensorFlow (\cite{tf}) backend. 
In addition, we used Kaldi (\cite{kaldi}), an open-source speech recognition toolkit, for speech signal processing such as i-vector (\cite{ivec}) extraction. 

\subsection{The effects of the adversarial processes}
In this subsection, we provide the empirical evidence of the assumption on the adversarial processes mentioned in Section 2 and Section 3. 
In this experiment, we remove the information of the digit identities of the images from the feature by the adversarial processes and observe the outputs of the digit recognizers relying on the feature to see if the information is really removed and the output is independent of the input.
To do so, we designed the experiments using the MNIST dataset (\cite{mnist}) and simple convolutional neural network (CNN) with five hidden layers and 10 output nodes. 
First, the CNN was trained to identify 10 classes in the MNIST based on CCE loss. 
After sufficient training of the CNN, the first three layers were trained again based on the conventional adversarial loss (defined as Eq. (3)) or the proposed one (defined as Eq. (4)). 
Then we evaluated the performances of the CNN using the test set of the MNIST to confirm the effects of the adversarial processes. 
For performance evaluation, we applied the argmin function or the argmax function to the output of the CNN to calculate the accuracy on identifying each image. 
Table 1 shows the performances depending on the adversarial processes. 
When the identification is performed with the argmax function, the system without the adversarial process has a high accuracy of 99\% or more, and the system with the conventional adversarial process has an accuracy of 0\%. 
The system with an accuracy of 0\% can be misunderstood as if it does not perform identification at all. 
However, assuming a system that does not actually perform the identification at all, the system should have 10\% accuracy for 10 classes. 
This is because the expected accuracy for a system that outputs an arbitrary value for every input is about 10\%. 
Therefore, the system with an accuracy of 0\% cannot be interpreted as not performing the identification but interpreted as performing the identification in other ways. 
The results of the experiments using the argmin function can be the evidence of this interpretation. 
When the identification is performed with the argmin function, the system without the adversarial process has an accuracy of 0\%, and the system with the conventional adversarial process has an accuracy of 96\%. 
These results show that the conventional adversarial process is not suitable for eliminating the subsidiary information because it only changes the way of identification instead of degrading the performance of the model. 
On the other hand, the system with the proposed adversarial process has accuracies of about 10\% regardless of the identification ways (argmin or argmax). 
Therefore, we confirmed that the proposed adversarial process can make the output independent of the input and can efficiently degrade the performance of the model.

\begin{table}[t]
\caption{Accuracies (\%) of digit recognition experiments using different adversarial processes.}
\label{result_table2}
\vskip 0.15in
\begin{center}
\begin{small}
\begin{tabular}{lccc}
\toprule
identification methods & w/o adversarial & Conventional adversarial & Proposed adversarial \\
\midrule
argmax & 99.37 & 0.00 & 10.24\\
argmin & 0.00 & 96.50 & 13.60\\

\bottomrule
\end{tabular}
\end{small}
\end{center}
\vskip -0.1in
\end{table}

\subsection{Eliminating channel information in speaker identification}
Here, we use the proposed CAN framework to reduce channel information in speaker identification task. 
Speaker identification is a task that identifies the person who spoke the input speech.
In a typical speaker identification task, it is assumed that only the speech of known speakers is inputted, thus limiting the candidate group to be identified.
Channel information is well-known as one of factors that decreases speaker identification performance (\cite{svm_channel}). 
Furthermore, channel information varies depending on the recording devices and transmission method. 
Therefore, even if utterances are from a known speaker, they can have different characteristics owing to the channel information. 
The purpose of applying the proposed CAN framework is to eliminate channel information.
Eliminating channel information from speech data is a challenging task, because there is no specific target sample which is a clean speech utterance without any channel information. 
For example, we can create noisy speech by inserting noise into clean speech; the clean and noisy speeches can be used as the target and source sample, respectively, for the task of eliminating noise from speech. 
Thus, if there are target and source samples, it is possible to train models, such as a stacked denoising autoencoder, to eliminate noise information. 
However, there is no target sample (clean speech) for the task of eliminating channel information, because, as previously mentioned, the characteristics of the recording device are inevitably reflected in the process of recording the speech.

The experiments for speaker identification were designed using the RSR2015 dataset (\cite{rsr}) as follows. 
Six mobile devices (five smartphones and one tablet) were used for collecting speech data for the RSR2015. 
Three devices were assigned to each speaker. 
The performance of speaker identification was evaluated for 143 female speakers among a total of 300 speakers. 
For training, 10 utterances for each speaker were used. 
Then, identification was performed using utterances that were about one second long. 
In addition, as input for the encoder, 200-dimensional i-vectors were extracted for each utterance. 
The i-vector is a technique that is used for representing utterances of varying lengths as a vector with fixed dimensionality (\cite{ivec}). 
In particular, the encoder E receives an i-vector and outputs the code. 
The two models (M and S) receive the code and perform speaker identification and channel identification, respectively. 

In order to validate the performance of the proposed CAN framework, we compared the performances of the four systems: one without an adversarial process, two with the conventional adversarial processes and the proposed process. 
We report performances of each system by following similar manner in \cite{cogan}; averaging error rates over 5 networks that were randomly selected after sufficient training. 
More details of the experiments are in Appendix-A.  

\begin{table}[t]
\caption{Experimental results for speaker identification using the three frameworks compared in our study.}
\label{result_table1}

\vskip 0.15in
\begin{center}
\begin{small}

\begin{tabular}{lcccr}
\toprule
 & w/o adversarial process & w/ \cite{yu2017adversarial} & w/ \cite{GRL2016} & w/ proposed CAN \\
\midrule
Error rate(\%) & 5.69 $\ \pm$ 0.13  & 5.53 $\ \pm$ 0.09 & 5.32 $\ \pm$ 0.09 & 4.88 $\ \pm$ 0.12 \\

\bottomrule
\end{tabular}

\end{small}
\end{center}
\vskip -0.1in
\end{table}

Table 2 lists the experimental results for speaker identification. 
The system without adversarial process means a network consisting of only E and M. 
The framework of \cite {yu2017adversarial} did not show a significant performance improvement over the system without the adversarial process.  
The GRL-based framework (\cite{GRL2016}) showed a relative error exclusion of about 6.5\% compared to the system without the adversarial process. 
In contrast, our proposed CAN framework showed the lowest error rate, with a relative error reduction of about 14.2\% compared to the GRL-based framework.

\begin{table*}[t]
\caption{Experimental results for digit recognition using different frameworks.}
\label{result_table2}
\vskip 0.15in
\begin{center}
\begin{small}
\begin{tabular}{lcccr}
\toprule
Error rate(\%) &\begin{tabular}{@{}c@{}}Baseline \\ (src only train)\end{tabular}  &\begin{tabular}{@{}c@{}}ADDA \\  \cite{ADDA}\end{tabular} &\begin{tabular}{@{}c@{}}ADDA \\ (our implementation)\end{tabular}& Proposed CAN \\
\midrule
Target domain (USPS) & 17.59 $\ \pm$ 0.39  & 10.60 $\ \pm$ 0.20 & 8.51 $\ \pm$ 0.30 & 5.51 $\ \pm$ 0.33 \\
Source domain (MNIST)& 2.62 $\ \pm$ 0.23  & - & - & 2.50 $\ \pm$ 0.23 \\
\midrule
Target domain (MNIST) & 30.76 $\ \pm$ 1.30  & 24.00 $\ \pm$ 1.80 & 19.69 $\ \pm$ 1.21 & 16.74 $\ \pm$ 1.03 \\
Source domain (SVHN)& 5.70 $\ \pm$ 0.16  & - & - & 5.40 $\ \pm$ 0.15 \\
\bottomrule
\end{tabular}
\end{small}
\end{center}
\vskip -0.1in
\end{table*}

\subsection{Unsupervised domain adaptation}

Unsupervised domain adaptation is a task that makes a classifier suitable for a target domain. 
The main challenge of this task is that only unlabeled data are provided from the target domain. 
Though both images and labels were used to train the source domain model, only the images were used for adaptation to the target domain. 
Therefore, the classifier, which is trained using labeled data of the source domain, should be generalized to the target domain.
However, because of the phenomenon of dataset bias, a classifier from one domain does not generalize well to another domain (\cite{dataset_biase}). 
We assume that the dataset bias is caused by domain information included in the data. Considering this assumption, the domain information represents characteristics of each dataset. Therefore, we performed the unsupervised domain adaptation task using the proposed CAN framework to eliminate domain information.

We designed the unsupervised domain adaptation task using the MNIST (\cite{mnist}), USPS, and SVHN (\cite{svhn}) datasets. 
Several examples of the MNIST, USPS, and SVHN datasets dataset are depicted in Fig. 4. 
\begin{figure}
\begin{center}
\centerline{\includegraphics[width=0.5\columnwidth, trim=4 4 4 4,clip]{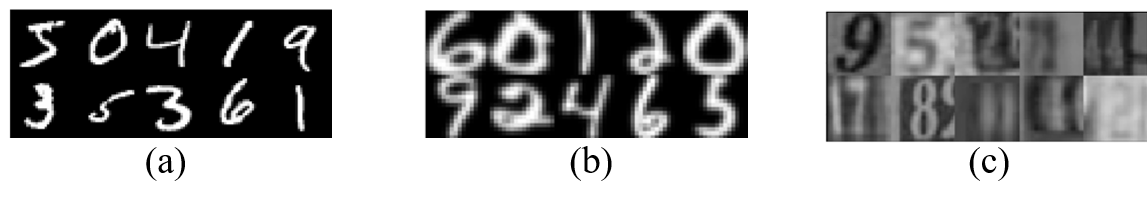}}
\caption{Examples from the (a) MNIST, (b) USPS and (c) SVHN datasets.}
\label{fig1}
\end{center}
\vskip -0.2in
\end{figure}
For this performance comparisons, we conducted two kinds of experiments on domain adaptation. 
As the first experiment, we adopted the experimental protocol described in (\cite{ADDA}). 
We used the MNIST dataset as the source domain, whereas the USPS dataset as the target domain for domain adaptation. 
We additionally conducted domain adaptation experiment using different, larger dataset, SVHN instead of USPS with MNIST dataset. 
In this experiment, the SVHN dataset is used as the source domain, and the MNIST dataset is used as the target domain. 
Such setting was introduced in \cite{ganin2016domain}, as adaptation between domains which are significantly different in appearance.
Unlike previous studies that used LeNet (\cite{lenet}) for digit recognition, we used a model based on the maximum feature map (MFM) (\cite{LCNN}). 
More details of the experiments are in Appendix-B.  

We compared the performances of the proposed framework with the adversarial discriminative domain adaptation (ADDA, \cite{ADDA}). 
\cite{ADDA} summarized various adversarial processes that can be used for domain adaptation and proposed a GAN loss-based adaptation framework. 
In order to validate the performance of the proposed framework, we compared the performances of the system without the adaptation, with ADDA, with the proposed process, along with the performance reported in \cite{ADDA}.

Table 3 shows the error rates of different frameworks on domain adaptation. 
We confirmed that the performance of the ADDA implemented by us is higher than the previously reported performance of ADDA (\cite{ADDA}). 
From this result, we found that the MFM-based model is more suitable for the domain adaptation task than the one using LeNet even if the size of the MFM-based model is smaller than the LeNet.
Our proposed CAN framework showed the lowest error rate with a relative error reduction about 35\% compared to the ADDA framework in the experiments on adaptation to USPS from MNIST.
The proposed framework does not generate a separate encoder for the target domain data during the domain adaptation process. 
Therefore, it is possible to recognize the source domain data using the model adapted to the target domain. 
In contrast, the ADDA generates an additional encoder for the target domain; therefore, the adapted model cannot recognize data from the source domain. 
Using the same model for the target and source domains has advantages, including reducing the size of the model and increasing its availability. 
However, model optimization using conventional methods might be performed poorly because the same model must process images from two separate domains (\cite{ADDA}). 
The results obtained for the performance of the proposed framework in the case of the source and target domains are contrary to this conventional knowledge. 
In particular, the proposed CAN framework maintained the performance on the source domain and successfully adapted the model to the target domain simultaneously. 
We could confirm the similar trend in the experiments on adaptation to MNIST from SVHN. 
Our proposed CAN framework showed the lowest error rate with a relative error reduction about 15\% compared to the ADDA framework in target domain. 
In addition, we also confirmed that the error rate of the source domain is reduced by about 5\% through the process of eliminating domain information.

\subsection{Analysis of application results of the adversarial processes}

\begin{figure}[ht]
\begin{center}
\centerline{\includegraphics[width=1.\columnwidth,trim=4 4 4 4,clip]{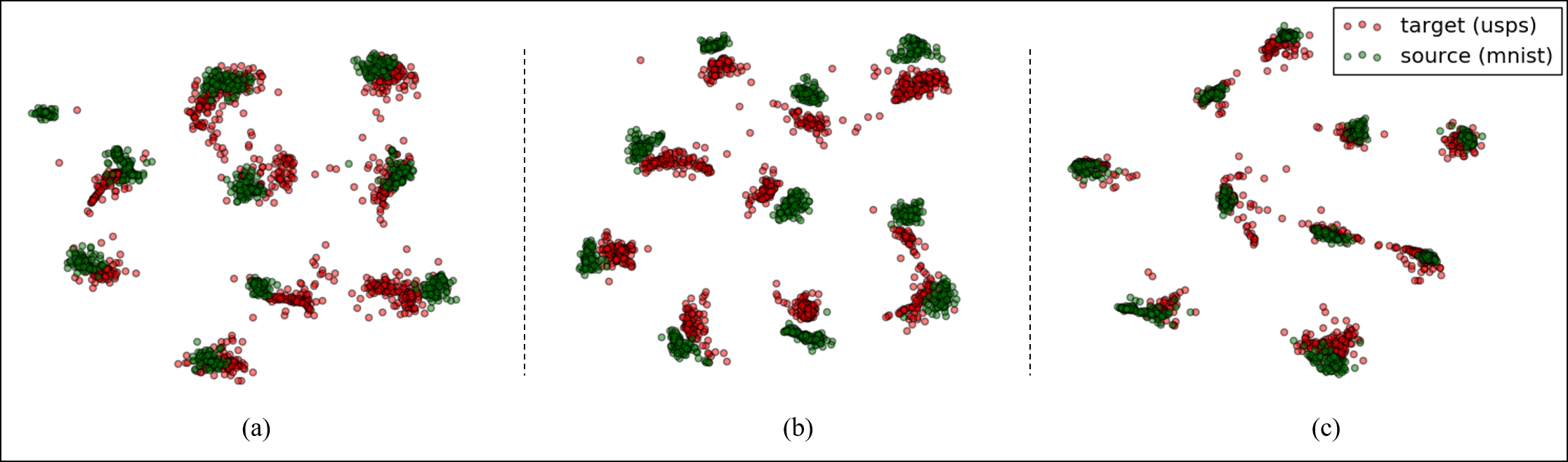}}
\caption{Visualization of codes from various encoders (colors indicate different domains). (a) No adaptation, (b) Adaptation based on ADDA, (c) Adaptation based on the proposed CAN framework.}
\label{fig1}
\end{center}
\vskip -0.2in
\end{figure}
The proposed CAN focuses on eliminating subsidiary information by using cosine similarity-based loss, while the conventional adversarial methods focus on confusing S. 
However, in the previous two experiments, only the performance improvements for the primary tasks (speaker identification and digit recognition) were reported, so it is difficult to confirm whether or not the subsidiary information was effectively eliminated by the proposed CAN framework. 
Therefore, in this subsection, we analyze the outputs of E trained by the adversarial processes to confirm that the subsidiary information is actually eliminated. 
To do so, we analysed the outputs of E in two ways; the first one is visualizing the outputs of E, and the second one is identifying subsidiary information using the outputs of E. 

Fig. 5 shows the two-dimensional codes from various Es considered in the experiment on domain adaptation. 
The codes are extracted from the training data of the MNIST and USPS datasets, and visualized using the t-SNE technique (\cite{tsne}). 
The target and source domain codes are indicated using red and green dots, respectively. 
The visualization result of the baseline (Fig. 5-(a)) shows that the codes from the source domain are represented by 10 clusters (referred to as 10 classes) while the codes from the target domain are represented by 9 clusters. 
This phenomenon considerably degrades the performance of the model M on the target domain even if the distributions of the source and target domains seem similar. 
The result of the ADDA system (Fig. 5-(b)) shows that the codes of the target domain are expressed in 10 clusters, thus the model is well adapted to the target domain. 
However, it was observed that there is a difference in the distributions of the codes from the source and target domains because the domain information is still retained in the codes. 
In contrast, considering the result of our proposed CAN framework (Fig. 5-(c)), it can be observed that the distribution of the source and target domains almost overlap. 

\begin{figure}[ht]
\begin{center}
\centerline{\includegraphics[width=0.8\columnwidth]{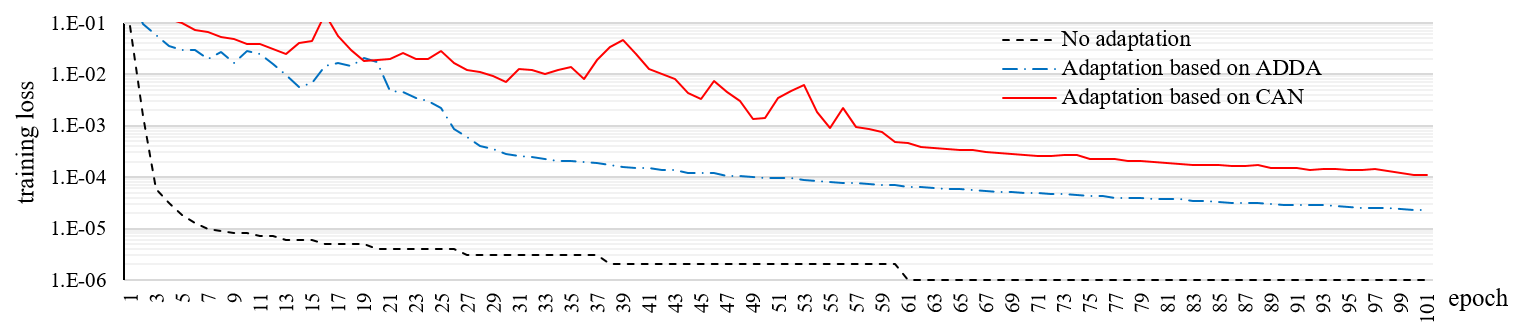}}
\caption{Experimental results for domain recognition using codes from E. The results show that it was harder to recognize the domains of codes from E adapted by the proposed method than other cases.}
\label{fig1}
\end{center}
\vskip -0.2in
\end{figure}

We demonstrate the effect of the proposed method through an experiment to recognize the subsidiary information of the codes from E. 
Here, recognizing the subsidiary information means identifying the domains of codes (MNIST or USPS).
For this purpose, multiple Es from the experiments on adaptation to USPS from MNIST in Section 4.3 were used; Es were adapted using CAN or ADDA, or not adapted. 
We observed the training loss while training a new model $\textrm{S}_2$ to identify the domain of the codes (see Fig. 6). 
The model $\textrm{S}_2$ has the same structure with S and performs the same operation, but the parameters were re-initialized and trained.
We assumed that the training losses of $\textrm{S}_2$ can reflect the amount of subsidiary information remaining in the codes. 
Fig. 6 shows that recognizing the domains of codes from E adapted by the proposed method is harder than other cases. 
Therefore, we concluded that using the proposed CAN is more appropriate for eliminating the subsidiary information and domain adaptation, than the conventional methods which confuses S.

\section{Conclusions}
In the DAN framework, the adversarial process is applied to eliminate the subsidiary information that can degrade the performance of the primary task. 
For this purpose, after training S, which recognizes the subsidiary information, we apply the adversarial process to degrade the performance of S. 
However, we reveal that the conventional adversarial process has a limitation. 
In particular, it is shown that the conventional adversarial process, maximizing CCE, is not suitable to degrade the performance of the model. 
To overcome this limitation of the conventional adversarial process, we proposed the novel adversarial process based on cosine similarity (CAN).
The proposed CAN framework degrades the performance of S by applying the concept of rotation to the codes. 

In the experiment for simulation of the adversarial processes, we confirmed that the proposed CAN effectively degraded the performance of S while the conventional adversarial process just confused S (see Section 4.1).
The experimental results in Section 4.2 and Section 4.3 show that the proposed CAN framework can be effectively applied to either speaker recognition or domain adaptation. 
Finally, based on the analysis in Section 4.4, we concluded that the proposed CAN effectively eliminate the subsidiary information.

\bibliography{example_paper}
\bibliographystyle{icml2019}

\end{document}